\documentclass[12pt]{article}
\usepackage{amsmath}
\usepackage{times}
\usepackage{graphicx}
\usepackage{color}
\usepackage{multirow}
\usepackage{rotating}
\usepackage{bbm}
\usepackage{latexsym}

\usepackage{comment}

\usepackage{doi}
\usepackage{hyperref}

\providecommand\doi[1]{\href{https://doi.org/#1}{\url{#1}}}

\usepackage{amsthm}
\usepackage{amsfonts}
\usepackage{amssymb} 
\usepackage{bbm} 

\usepackage{amsmath} 

\newcommand\numberthis{\addtocounter{equation}{1}\tag{\theequation}}

\usepackage{booktabs}
\usepackage{array}
\usepackage{longtable}
\usepackage{hyperref}

\DeclareMathOperator*{\argmin}{arg\,min}

\DeclareMathOperator*{\trace}{tr}

\DeclareMathOperator*{\vect}{vec}

\usepackage[ruled,vlined]{algorithm2e}

\numberwithin{equation}{section}

\newtheorem{theorem}{Theorem}

\usepackage{xcolor}

\newcommand{\captionfonts}{\normalsize}

\newtheorem{lemma}{Lemma}

\makeatletter  
\long\def\@makecaption#1#2{%
  \vskip\abovecaptionskip
  \sbox\@tempboxa{{\captionfonts #1: #2}}%
  \ifdim \wd\@tempboxa >\hsize
    {\captionfonts #1: #2\par}
  \else
    \hbox to\hsize{\hfil\box\@tempboxa\hfil}%
  \fi
  \vskip\belowcaptionskip}
\makeatother   

\begin{document}
\hspace{13.9cm}1

\ \vspace{20mm}\\

{\LARGE Gauge-optimal approximate learning for small data classification problems}

\ \\
{\bf \large Edoardo Vecchi$^{\displaystyle 1}$, Davide Bassetti$^{\displaystyle 2}$, Fabio Graziato$^{\displaystyle 3}$, Luk\'{a}\v{s} Posp\'{i}\v{s}il$^{\displaystyle 4}$ and Illia Horenko$^{\displaystyle 2}$}\\
{$^{\displaystyle 1}$Universit\`{a} della Svizzera Italiana (USI), Faculty of Informatics, Institute of Computing, Via la Santa 1, 6962 Lugano, Switzerland}\\
{$^{\displaystyle 2}$ Technical University of Rhinland-Palatinate  Kaiserslautern-Landau (RPTU), Faculty of Mathematics, Chair for Mathematics of AI, 67663 Kaiserslautern, Germany}\\
{$^{\displaystyle 3}$ Independent researcher, Via IV Novembre 244/115, 22070 Valmorea (CO), Italy}\\
{$^{\displaystyle 4}$ VSB Ostrava, Department of Mathematics, Ludvika Podeste 1875/17 708 33 Ostrava, Czech Republic}\\


\thispagestyle{empty}
\markboth{}{NC instructions}
\ \vspace{-0mm}\\
\newpage
\begin{center} {\bf Abstract} \end{center}
Small data learning problems are characterized by a significant discrepancy between the limited amount of response variable observations and the large feature space dimension. In this setting, the common learning tools struggle to identify the features important for the classification task from those that bear no relevant information, and cannot derive an appropriate learning rule which allows to discriminate between different classes. As a potential solution to this problem, here we exploit the idea of reducing and rotating the feature space in a lower-dimensional gauge and propose the Gauge-Optimal Approximate Learning (GOAL) algorithm, which provides an analytically tractable joint solution to the dimension reduction, feature segmentation and classification problems for small data learning problems. We prove that the optimal solution of the GOAL algorithm consists in piecewise-linear functions in the Euclidean space, and that it can be approximated through a monotonically convergent algorithm which presents---under the assumption of a discrete segmentation of the feature space---a closed-form solution for each optimization substep and an overall linear iteration cost scaling. The GOAL algorithm has been compared to other state-of-the-art machine learning (ML) tools on both synthetic data and challenging real-world applications from climate science and bioinformatics (i.e., prediction of the El Ni\~no Southern Oscillation and inference of epigenetically-induced gene-activity networks from limited experimental data). The experimental results show that the proposed algorithm outperforms the reported best competitors for these problems both in learning performance and computational cost.

\section{Introduction}\label{sec:introduction}

While the amount of available information---along with the challenges it bears---grows exponentially in the era of Big Data \cite{sagiroglu2013big, fan2014challenges}, a plethora of application domains is still forced to operate in the small data regime \cite{qi2020small, konietschke2021small}, i.e., a scenario in which the size $T$ of the available training set statistics is relatively small when compared to the large dimension $D$ of the feature space. Among the others, we can mention, e.g.,  financial asset management \cite{israel2020can}, medical imaging \cite{fakoor2013using, zhang2019survey, oh2020deep, tartaglione2020unveiling}, climate science \cite{knusel2019applying}, oceanography \cite{jin2017deep}, biology \cite{walsh2021dome}, oncology \cite{singh2015feature, basavegowda2020deep}, epidemiology \cite{zeroual2020deep} and psychiatry \cite{bzdok2018machine, koppe2021deep}. However, despite the growing relevance of small data learning problems, the literature still lacks a common theoretical framework: the definition of ``small data'' itself is strictly related to the characteristics of the problem under consideration \cite{donoho1995noising, horenko2020scalable, vecchi2022espa+}, and even the terminology used widely differs depending on the specific application domain. In particular---while some studies have highlighted the problems stemming from different naming conventions \cite{DUINTJERTEBBENS2007423}---systematic work in this specific direction still needs to be done, and currently we are forced to move between variants of the ``few observations, many variables'' sentence (like, e.g., high-dimension small-sample size data) and terms like ``fat big data'', ``short data'', and ``wide data''. For the sake of clarity, in this letter we say that a certain classification problem pertains to the ``small data  learning regime'' if the number $T$ of samples (or observations) of the data statistics is smaller than the number $D$ of available explanatory features.

Beyond the theoretical issues outlined above, the key point that we want to stress is that we are still in dire need of machine learning (ML) and deep learning (DL) methods  capable of operating in the small data regime. As outlined in previous studies pertaining to different application domains \cite{keshari2020unravelling, d2020structural, doi:10.1177/00368504211029777}, current state-of-the-art ML approaches tend to show a certain lack of robustness when dealing with a limited amount of data, thus achieving a poor performance on the test set due to the impossibility of generalizing what has been learned in the training set \cite{rice2020overfitting, hosseini2020tried, Montesinos2022}. This situation is further complicated by the fact that the apparently satisfying performance achieved by ML in several studies could potentially represent a biased result, induced by the very same limited size of the training set and by an insufficient control for overfitting \cite{10.1371/journal.pone.0224365}, thus opening an additional discussion---which is beyond the scope of this paper---about which metrics would be best suited to assess the performance of ML algorithms in the small data regime. While some approaches like data augmentation or transfer learning have been successfully applied in some specific domains, their applicability is usually limited in the natural sciences, due to the violation of their intrinsic assumptions \cite{vecchi2022espa+}. In order to address this issue, recent studies have been focused either on benchmarking the currently available methods in specific small data applications \cite{doi:10.1021/acs.chemrev.3c00189, vecchi2023entropic} or on proposing methods tailored to the solution of certain classes of small data problems---like, e.g., ultra-low-radiation medical image denoising \cite{horenko2022lowcost}, binary classification \cite{vecchi2022espa+} or linear and nonlinear regression problems \cite{horenko2023oncheap}.

However---despite the reported promising advances---further research is still needed: the proposed methods need to be thoroughly tested in many different settings before becoming state-of-the-art tools, while methods capable of addressing further layers of complexity---like, e.g., concept drift or a strong imbalance in a small data setting---are yet to be proposed. In particular, imbalanced classification tasks are particularly challenging in the small data learning regime, due to the fact that many of the standard approaches are usually impractical in this setting. Among the most widely used techniques for imbalanced classification, we can find undersampling and oversampling \cite{drummond2003c4, yap2014application, kaur2018comparing, mohammed2020machine}, but both these methods are actually detrimental to the classifier performance in presence of few data instances and many features. Indeed---on the one hand---undersampling tries to close the gap between the majority and the minority class by systematically dismissing observations from the former, so that we have a similar number of instances in both groups. However, in the small data regime the quantity of observations is already limited, and this approach tends to reduce even further the amount of information that can be extracted from the training data to perform the classification task. On the other hand, oversampling methods---like, e.g., the Synthetic Minority Over-sampling Technique (SMOTE) \cite{chawla2002smote}---tackle the data imbalance by generating artificial copies of the minority class, thus presenting the same limitations previously discussed for other data augmentation approaches. In particular, given the small number of instances in the minority class, using this technique could either produce biased observations that further complicate the classification task, or could lead to a potential overfitting of the minority class \cite{fernandez2018learning}.

In this letter---in order to further contribute to the group of methods dealing with learning tasks pertaining to the small data regime---we exploit the idea of optimal gauge-rotation \cite{NIPS2009_82cec960} to propose an algorithm that provides a joint solution to the dimension reduction, feature segmentation and classification learning problems. In a nutshell, the obtained Gauge-Optimal Approximate Learning (GOAL) algorithm---after finding the optimal rotation of the input data in a lower-dimensional gauge---seeks for the prediction-optimal box discretization which minimizes the Kullback-Leibler divergence between the true labels and the predicted labels. By imposing certain assumptions---which are further detailed in the methodological part---about the discretization of the data space, we prove that GOAL results in a monotonically convergent iterative algorithm with analytically solvable substeps and a linear scaling of the overall computational cost. This last point is particularly important, since also the GOAL method---as the other ML and DL algorithms---relies on a set of hyperparameters that needs to be tuned during its training. In general, choosing the correct set of hyperparameters can have a massive impact on the performance of ML models, and the development of strategies aimed at automatically selecting the best parameter values is a recurrent object of analysis \cite{kohavi1995automatic, hutter2015beyond, luo2016review}. However, due to the increasing complexity of ML and DL algorithms, hyperparameter optimization approaches are faced with several challenges---like, e.g., computationally expensive function evaluations, high-dimensional hyperparameter grids and a generalized uncertainty about the range within which the optimal parameter value should be searched for  \cite{feurer2019hyperparameter}. In particular, the aforementioned range can change significantly with respect to the specific application domain, and the limited training size of small data problems hinders the possibility of inferring optimal intervals to investigate, often forcing us to perform a grid search \emph{ex novo} for every data set. Thus, it is extremely important to develop methods that are not only well-performing, but also computationally efficient, since they would allow to address the hyperparameter optimization problem through an extensive and well-defined grid search, without being constrained by the number of features or data instances. This is the reason why---as we will explain in Section \ref{sec:GOAL}---we decided to trade some of the flexibility in the input data discretization for a massive gain in computational cost, thus moving from a polynomial to a linear complexity. Finally---in order to provide a first assessment of GOAL performance---we consider its application to both synthetic data and to real-world data sets dealing with climate time series prediction and small experimental omics data in the area of bioinformatics. Through a thorough benchmarking with several other ML and DL methods, we show that in some instances---e.g., 24-month-ahead-prediction of the El Ni\~{n}o Southern Oscillation phenomenon or inference of epigenetically-induced gene-activity networks---the GOAL algorithm outperforms its competitors in terms of predictive performance, while its computational cost remains lower than the majority of the other algorithms. On the other hand, the binary classification of synthetic data from bioinformatics shows that  the GOAL algorithm---while performing better than other classes of ML methods---cannot outperform the recently proposed advanced entropic Scalable Probabilistic Approximation (eSPA+) algorithm \cite{vecchi2022espa+}, thus hinting towards further research in this class of problems.

The rest of this letter is organized as follows. In section \ref{sec:GOAL} we derive the GOAL algorithm and prove some results concerning the existence of a closed-form solution for each minimization substep and the linear scaling of the computational cost. In section \ref{sec:experiments}, we briefly explain the data set used in the experiments and discuss the results of the benchmarking of the GOAL algorithm with its main state-of-the-art competitors from ML and DL. Section \ref{sec:conclusions} briefly wraps up the letter, by presenting the open challenges and some potential promising directions for future research.

\section{Material and Methods: the GOAL algorithm}\label{sec:GOAL}

In this section, we present the derivation of the general formulation of the GOAL method, and then prove some theoretical results about the monotone convergence of the algorithm and the existence of a closed-form solution under a discrete segmentation assumption, which allows to achieve an overall linear iteration cost scaling. 

\subsection{Mathematical Derivation of GOAL}

\begin{figure}[h!]
\begin{center}  
    \includegraphics[width=\textwidth]{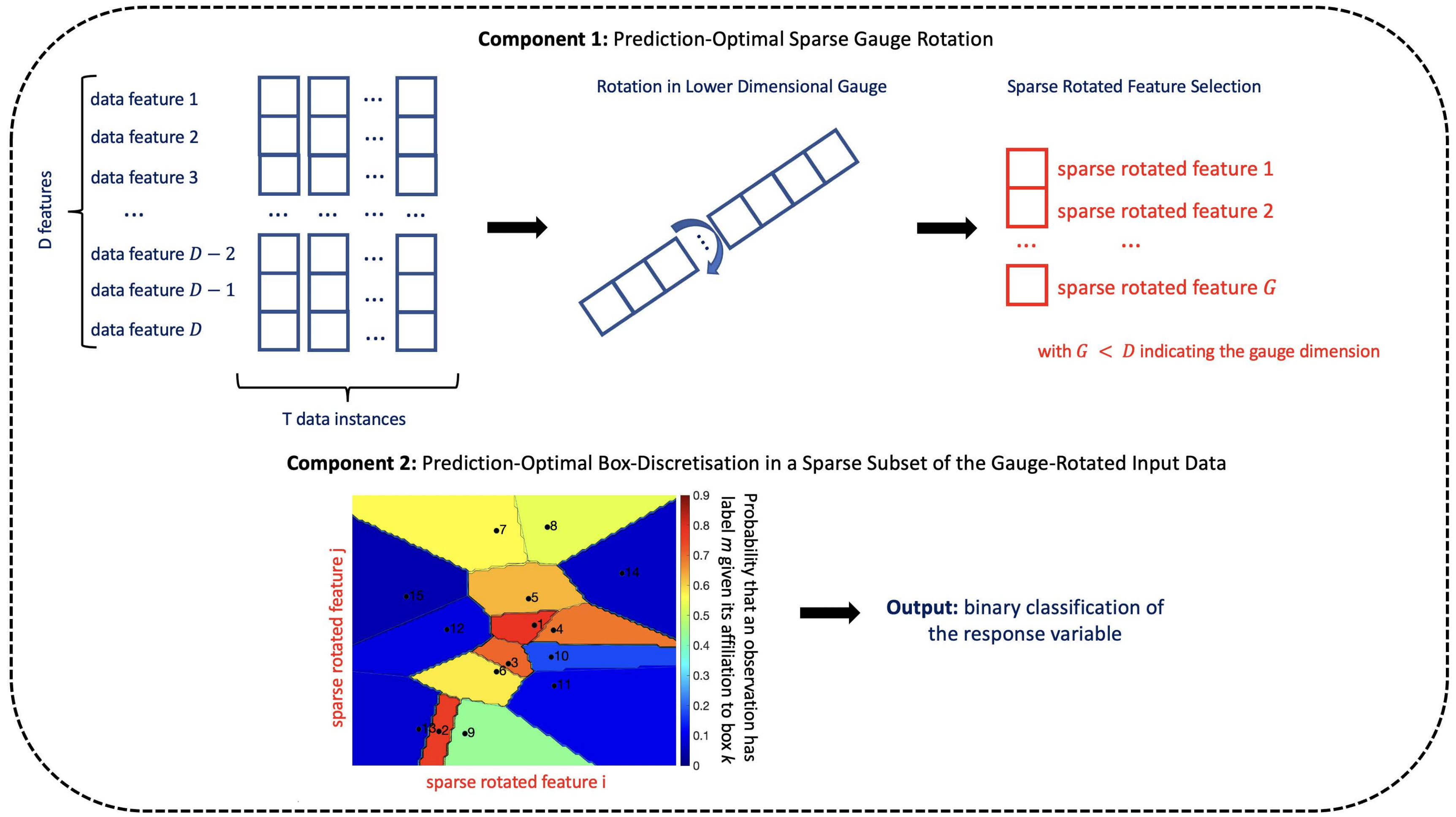}  
 \end{center}
   \caption{\emph{Overview of GOAL main components: after finding the optimal rotation matrix in a lower dimensional gauge, we search for the discretization of the reduced features which minimizes the Kullback-Leibler distance from the true label probabilities.}}
    \label{fig:goal_overview}
\end{figure}

While partially combining the ideas about data space discretization of the more general Scalable Probabilistic Approximation (SPA) framework \cite{gerber20} and about supervised classification reported in \cite{horenko2020scalable, vecchi2022espa+}---the GOAL algorithm approaches the solution of binary classification problems from a different perspective. Specifically, its main intuition consists in rotating and reducing the coordinate system of the input data matrix $X\in\mathbb{R}^{D\times T}$, in order to find the best Euclidean discretization of the problem that is simultaneously optimal also in terms of Kullback-Leibler divergences between the discretization probabilities and the labels. The transformation of the original data space is operated by means of a rotation matrix $R\in\mathbb{R}^{D\times G}$, with $G$ representing a hyperparameter indicating the dimension of the lower dimensional gauge on which the data is projected. The GOAL algorithm then solves the classification problem by minimizing the regolarized functional:
\begin{equation}\label{eq:GOAL}
	 \mathcal{L}_{\text{GOAL}}=\frac{1}{T}\sum_{d=1}^D\sum_{t=1}^T\left(X_{d,t}-\left\{RS\Gamma\right\}_{d,t}\right)^2 -\frac{\varepsilon_{\text{CL}}}{TM}\sum_{m=1}^M\sum_{t=1}^T\Pi_{m,t}\log\left(\sum_{k=1}^{K}\Lambda_{m,k}\Gamma_{k,t}\right),\\
\end{equation}
under the condition that $\Gamma$, $\Lambda$ and $R$ belong to the following feasible sets:
\begin{align}
	\Omega_{\Gamma} &:= \left\{ \Gamma\in \mathbb{R}^{K \times T} \bigg | \, \forall k,t: \; \Gamma_{k,t} \in [0,1]\; \wedge \; \forall t: \; \sum_{k=1}^{K} \Gamma_{k,t}=1 \right\},\label{eq:Gamma}\\
	\Omega_{\Lambda} &:= \left\{ \Lambda\in \mathbb{R}^{M \times K} \bigg | \, \forall m,k: \; \Lambda_{m,k} \in [0,1]\; \wedge \; \forall k: \; \sum_{m=1}^M \Lambda_{m,k}=1 \right\},\label{eq:Lambda}\\
	\Omega_R &:= \left\{ R\in \mathbb{R}^{D\times G} \bigg | \, R^\intercal R = I_{G} \right\},\label{eq:R}
\end{align}
with $I_{G}$ indicating an identity matrix of dimension $G$. In the above formulation, matrix $S\in\mathbb{R}^{G\times K}$ contains the coordinates of the discrete, non-overlapping boxes which partition the space of the gauge-rotated and reduced input data. Matrix $\Gamma\in\mathbb{R}^{K\times T}$---on the other hand---contains the affiliation of every data instance to one or more discretization boxes. It is important to notice that---in the more general formulation of the GOAL optimization problem---the cluster affiliations in the gauge-reduced space are fuzzy: i.e., each input data point can be assigned with a different probability to different boxes. While---in principle---this kind of fuzzy affiliation is possible, the assumption that each data instance belongs to a single discretization box allows the GOAL algorithm to present a closed-form solution in every optimization subproblem, as proved in theorem \ref{the:GOAL}. The second part of the functional in equation \ref{eq:GOAL} is specifically aimed at addressing the supervised labelling learning problem---akin to what has been suggested by \cite{horenko2020scalable} and \cite{vecchi2022espa+}---by minimizing the Kullback-Leibler divergence between the true data label probabilities $\Pi\in\mathbb{R}^{M\times T}$ and the estimated labels, where $\Lambda\in\mathbb{R}^{M\times K}$ is a stochastic matrix containing the conditional probabilities that a data point has a certain label $m$, given the fact that it has been affiliated with the discrete box $k$. The importance of the supervised labelling problem on the overall optimization procedure is regulated through $\varepsilon_\text{CL}$, a positive regularization term. The constrained minimization of the GOAL functional in Eq.~\ref{eq:GOAL} is, thus, aimed at finding the rotation in the low-dimensional gauge of size $G$---with $G$ significantly smaller than the dimension of the feature space $D$---that provides the optimal discretization with respect to both the Euclidean cluster affiliation errors and the relative entropy between the true and estimated data label probabilities. In Figure~\ref{fig:goal_overview}, we provide an intuitive graphical representation of the two main components behind the GOAL algorithm.
\vskip 0.25in

The constrained minimization of the GOAL objective function in equation \ref{eq:GOAL}, under the feasibility constraints in equations \ref{eq:Gamma}, \ref{eq:Lambda} and \ref{eq:R}, is performed through the iterative solution of four subproblems---namely, the $S$-step, the $\Gamma$-step, the $\Lambda$-step and the $R$-step---aimed at finding the optimal rotation and discretization matrices for the given classification problem, while keeping all the other variables fixed. In the next section, we provide several details about the algorithm convergence and complexity, and show that all optimization substeps admit a closed-form solution in case of discrete segmentation (i.e., when each point belongs to a single box or $\forall k,t: \, \Gamma_{k,t} \in \{0,1\}$).


\subsection{Theoretical Results about GOAL Solution and Convergence}

Here, we want to provide several details about the GOAL algorithm and its substeps. First of all, we want to show that the $R$-step---which is aimed at finding the optimal rotation matrix in the lower-dimensional gauge---can be analytically computed starting from the general problem formulation in equation \ref{eq:GOAL}. The results are proved in the following lemma \ref{lemma:optimal_rotation}, which considers only the first term of the GOAL objective function $\mathcal{L}_\text{GOAL}$, since the relative entropy term is constant with respect to the optimal rotation.

\begin{lemma}\label{lemma:optimal_rotation}
	The constrained minimization problem	
	\begin{equation}\label{eq:GOALlemma}
	\begin{aligned}
		 \argmin_{R\in\Omega_R} \quad & f(R) = \left\|X - RS\Gamma \right\|^2_F = \sum_{d=1}^D\sum_{t=1}^T\left(X_{d,t}-\left\{RS\Gamma\right\}_{d,t}\right)^2,\\
		\text{with} \quad  & \Omega_R := \left\{ R\in \mathbb{R}^{D\times G} \bigg | \, R^\intercal R = I_G \right\},
	\end{aligned}
	\end{equation}
	is solved by an analytically computable optimal rotation matrix $R^*$, defined as:
	\begin{equation}
		R^* = U\tilde{I}V^\intercal.
	\end{equation}
\end{lemma}

\begin{proof}
	To find the optimal rotation matrix $\displaystyle R^*\in\mathbb{R}^{D\times G}$, we need to minimize the following quadratic functional:
	\begin{equation*}
		f(R) = \left\|X - RS\Gamma \right\|^2_F = X^{\intercal}X - 2X^{\intercal}RS\Gamma + \Gamma^{\intercal}S^{\intercal}S\Gamma,
	\end{equation*}
	where the term $\Gamma^{\intercal}S^{\intercal}R^{\intercal}RS\Gamma$ has been simplified since---by definition---$R^{\intercal}R=I_G$, with $I_G\in\mathbb{R}^{G\times G}$ representing an identity matrix. By including the constraint enforcing the orthogonality of the rotation matrix $R$ (i.e., a rectangular matrix with orthonormal columns), we obtain the following Lagrangian equation:	
	\begin{equation*}
		\mathcal{L}(R,\lambda) = X^{\intercal}X - 2X^{\intercal}RS\Gamma + \Gamma^{\intercal}S^{\intercal}S\Gamma + 2\sum_{i=1}^G\sum_{j=1}^G \lambda_{i,j} \left\{R^{\intercal}R - I_G\right\}_{i,j}.
	\end{equation*}
	By indicating with $\lambda\in\mathbb{R}^{G\times G}$ the matrix of Lagrange multipliers, we can compute the components of the gradient of $\mathcal{L}(R,\lambda)$ as follows:
	\begin{align*}
		\frac{\partial \mathcal{L}(R,\lambda)}{\partial R} &= -2X\Gamma^{\intercal}S^{\intercal} + 2R(\lambda + \lambda^{\intercal}),\\
		\frac{\partial \mathcal{L}(R,\lambda)}{\partial \lambda} &= R^{\intercal}R-I_G,
	\end{align*}
	and obtain the following set of KKT conditions \cite{nocedal06} for the original problem:
	\begin{align}
		R(\lambda+\lambda^{\intercal}) &= X\Gamma^{\intercal}S^{\intercal}, \label{eq:KKT1}\\
		R^{\intercal}R &= I_G. \label{eq:KKT2}
	\end{align}
	By multiplying both sides of equation \ref{eq:KKT2} from the left and from the right for the symmetric matrix $(\lambda+\lambda^{\intercal})$, we obtain that:
	\begin{equation*}
		(\lambda+\lambda^{\intercal})R^{\intercal}R(\lambda+\lambda^{\intercal}) = (\lambda+\lambda^{\intercal})^2,
	\end{equation*}
	which leads, by inserting the results of equation \ref{eq:KKT1} in the left side, to:	
	\begin{equation}\label{eq:afterKKT}
		S\Gamma X^{\intercal} X \Gamma^{\intercal}S^{\intercal} = (\lambda+\lambda^{\intercal})^2.
	\end{equation}
	If we then consider the generic singular value decomposition \cite{golub1965singular}:
	\begin{equation}\label{eq:svd}
		X \Gamma^{\intercal}S^{\intercal} = U\Sigma V^{\intercal},
	\end{equation}
	we can rewrite the symmetric positive semidefinite left term of equation \ref{eq:afterKKT} as:
	\begin{equation}\label{eq:svd2}
		S\Gamma X^{\intercal} X \Gamma^{\intercal}S^{\intercal} = V\Sigma^{\intercal}U^{\intercal}U\Sigma V^{\intercal} = VAV^{\intercal},
	\end{equation}
	where $\displaystyle A = \Sigma^{\intercal}\Sigma$ represents a diagonal matrix and $\displaystyle U^{\intercal}U$ has been omitted since it is an identity matrix. By combining the results in equations \ref{eq:afterKKT} and \ref{eq:svd2}, we have that:
	\begin{equation*}
		(\lambda+\lambda^{\intercal})^2 = VAV^{\intercal},
	\end{equation*}
	which can be equivalently rewritten as:
	\begin{equation}\label{eq:sqrt}
		\lambda+\lambda^{\intercal} = V\sqrt{A}V^{\intercal},
	\end{equation}
	with the following relationship between the diagonal elements of $A$ and $\Sigma$:
	\begin{equation}\label{eq:asigma_rel}
		\{\sqrt{A}\}_{ii} = \Sigma_{ii},\;\forall i =1,\dots,G.
	\end{equation}
	By taking the inverse of equation \ref{eq:KKT1} and by substituting \ref{eq:svd} and \ref{eq:sqrt} we obtain:
	\begin{equation}\label{eq:rstar}
		R^* = U\Sigma V^{\intercal}V (\sqrt{A})^{-1}V^{\intercal} = U\tilde{I}V^{\intercal},
	\end{equation}
	where---according to the results in equation \ref{eq:asigma_rel}---$\displaystyle \tilde{I}\in\mathbb{R}^{D\times G}$ is a rectangular diagonal matrix with an identity in the upper part and $V^\intercal V$ has been omitted since it is an identity matrix by construction of the singular value decomposition. It can be easily verified that $\displaystyle R^*$ in equation \ref{eq:rstar} satisfies the KKT conditions in \ref{eq:KKT1} and \ref{eq:KKT2}, and, thus, represents the unique minimizer of the constrained optimization problem in equation \ref{eq:GOALlemma}.
\end{proof}

The results presented in lemma \ref{lemma:optimal_rotation} allow us to find a closed-form solution for the computation of the optimal reduced-gauge-rotation matrix of the GOAL problem, since---as explained above---the constrained minimization problem in equation \ref{eq:GOALlemma} corresponds to the optimization of the objective function $\mathcal{L}_\text{GOAL}$ in equation \ref{eq:GOAL} for fixed values of the discretization parameters and conditional probability matrix. Furthermore, the optimal rotation matrix $R^*$ does not rely on any assumption on the segmentation of the original data space and, thus, can be applied both in case of a fuzzy or discrete cluster affiliation (i.e., when $\forall k,t$, $\Gamma_{k,t}\in [0,1]$ or $\Gamma_{k,t}\in \{0,1\}$ respectively). The solution of all other optimization subproblems is---on the other hand---heavily dependent on the nature of the chosen segmentation $\Gamma$ and can be handled by using, e.g., an interior-point method \cite{nocedal06}. In particular, in case of the general GOAL formulation reported in equation \ref{eq:GOAL}---which considers a fuzzy discretization of the input data space---the $\Gamma$-step represents the most computationally expensive step of the GOAL algorithm, with a complexity of $\mathcal{O}(TK^3 + KT(M+D))$, as outlined in \cite{horenko2020scalable} for the solution of a similar problem. Actually, the computational complexity of the $\Gamma$-step can be partially reduced by considering an upper bound of the optimal solution $\Gamma^*$, which can be obtained by solving an approximated convex quadratic programming problem, as shown in the next lemma \ref{lemma:optimal_gamma_qp}.

\begin{lemma}\label{lemma:optimal_gamma_qp}
	The optimal $\Gamma^*$ solving the constrained minimization problem
	\begin{equation}\label{eq:Gammalemma}
	\begin{aligned}
		 \argmin_{\Gamma\in\Omega_\Gamma} \quad & f(\Gamma) = \left\|X - RS\Gamma \right\|^2_F - \alpha \trace \left(\log(\Lambda\Gamma)\Pi^\intercal\right),\\
		\text{with} \quad  & \Omega_{\Gamma} := \left\{ \Gamma\in \mathbb{R}^{K \times T} \bigg | \, \forall k,t: \; \Gamma_{k,t} \in [0,1]\; \wedge \; \forall t: \; \sum_{k=1}^{K} \Gamma_{k,t}=1 \right\},
	\end{aligned}
	\end{equation}
	with $\alpha\in\mathbb{R}$ representing a strictly positive constant, is bounded from above by $\overline{\Gamma^*}$, which is the optimal solution of the convex quadratic programming (QP) problem
	\begin{equation}\label{eq:GammaQP}
		\argmin_{\Gamma\in\Omega_\Gamma} \;\; \frac{1}{2}\vect(\Gamma)^\intercal (I_T\otimes S^\intercal S) \vect(\Gamma) - \vect\left(S^\intercal R^\intercal X - \frac{\alpha}{2}\log(\Lambda)^\intercal \Pi\right)^\intercal \vect(\Gamma),
	\end{equation}
	under the same set of feasibility constraints of problem \ref{eq:Gammalemma}.
\end{lemma}

\begin{proof}
	In order to derive an upper bound of the optimal solution $\Gamma^*$, we can start by considering the alternative formulation of the second term of equation \ref{eq:Gammalemma}
	\begin{equation}\label{eq:secondTerm}
		-\alpha \trace \left(\log(\Lambda\Gamma)\Pi^\intercal\right) = - \alpha\sum_{m=1}^M\sum_{t=1}^T\Pi_{m,t}\log\left(\sum_{k=1}^{K}\Lambda_{m,k}\Gamma_{k,t}\right),
	\end{equation}
	and notice that, for fixed $m=\tilde{m}$ and $t=\tilde{t}$ and by dismissing the strictly positive constant $\alpha$, equation \ref{eq:secondTerm} results in the convex function:
	\begin{equation}\label{eq:psi1}
		\psi(\Lambda_{\tilde{m}}) = - \log\left(\sum_{k=1}^{K}\Lambda_{\tilde{m},k}\Gamma_{k,\tilde{t}}\right),
	\end{equation}
	where the arbitrary choice of considering $\Lambda_{\tilde{m}}$ as the independent variable is instrumental to the rest of the derivation. Equation \ref{eq:psi1} can be rewritten as:
	\begin{equation}\label{eq:psi2}
		\psi(\Lambda_{\tilde{m}}) = - \log\left(\frac{\sum\limits_{k=1}^{K}\Lambda_{\tilde{m},k}\Gamma_{k,\tilde{t}}}{\sum\limits_{k=1}^K\Gamma_{k,\tilde{t}}}\right),
	\end{equation}
	since, due to the feasibility constraint defined in problem \ref{eq:Gammalemma}, the probabilities of being associated to the different clusters sum up to 1. Since $\psi(\Lambda_{\tilde{m}})$ is a real convex function with positive coefficients $\Gamma_{1,\tilde{t}}, \dots, \Gamma_{K,\tilde{t}}$, we can apply Jensen's inequality \cite{jensen1906sur} to equation \ref{eq:psi2} and derive the following upper bound:
	\begin{equation}\label{eq:jensen_res}
		\psi(\Lambda_{\tilde{m}}) \leq - \frac{\sum\limits_{k=1}^K\Gamma_{k,\tilde{t}}\log(\Lambda_{\tilde{m},k})}{\sum\limits_{k=1}^K\Gamma_{k,\tilde{t}}} = -\sum\limits_{k=1}^K\Gamma_{k,\tilde{t}}\log(\Lambda_{\tilde{m},k}),
	\end{equation}
	By generalizing equation \ref{eq:jensen_res} to all values of $m$ and $t$, equation \ref{eq:Gammalemma} becomes:
	 \begin{equation}\label{eq:GammalemmaJensen}
		 \argmin_{\Gamma\in\Omega_\Gamma} \; f(\Gamma) \leq \argmin_{\Gamma\in\Omega_\Gamma} \left\|X - RS\Gamma \right\|^2_F - \alpha \trace \left(\log(\Lambda)\Gamma\Pi^\intercal\right).
	\end{equation}
	The upper bound in equation \ref{eq:GammalemmaJensen} can be rearranged by using the trace properties and by considering that---according to the definition in equation \ref{eq:R}---$R^\intercal R=I_G$:
	\begin{align*}
		&\trace\left((X-RS\Gamma)^\intercal(X-RS\Gamma)\right) - \alpha \trace \left(\log(\Lambda)\Gamma\Pi^\intercal\right)\\
		=&\trace(\Gamma^\intercal S^\intercal S\Gamma) - 2\trace(X^\intercal RS\Gamma) - \alpha \left(\Pi^\intercal\log(\Lambda)\Gamma\right) + \trace(X^\intercal X). \numberthis \label{eq:trace_prop}
	\end{align*}
	We know that the vectorization operator $\vect()$, for a generic matrix $A\in\mathbb{R}^{M\times N}$, is defined as a linear mapping from $\mathbb{R}^{M\times N}$ to $\mathbb{R}^{MN\times 1}$, which stacks the columns of the matrix in a single vector. By dismissing the constant term $\trace(X^\intercal X)$, which is irrelevant for the minimization of $\Gamma$, and by applying the properties from \cite{petersen2008matrix}:
	\begin{align*}
		\trace(A^\intercal B) &= \vect(A)^\intercal \vect(B), \\
		\vect(ABC) &= (C^\intercal\otimes A) \vect(B),
	\end{align*}
	we can rewrite equation \ref{eq:trace_prop} as follows:
	\begin{align*}
		& \trace(\Gamma^\intercal S^\intercal S\Gamma) - 2\trace(X^\intercal RS\Gamma) - \alpha \left(\Pi^\intercal\log(\Lambda)\Gamma\right) \\
		= & \vect(\Gamma)^\intercal \vect(S^\intercal S\Gamma) -2 \vect(S^\intercal R^\intercal X)^\intercal \vect(\Gamma) - \alpha \vect(\log(\Lambda)^\intercal \Pi)^\intercal \vect(\Gamma)\\
		= & \frac{1}{2}\vect(\Gamma)^\intercal (I_T\otimes S^\intercal S) \vect(\Gamma) - \vect\left(S^\intercal R^\intercal X - \frac{\alpha}{2}\log(\Lambda)^\intercal \Pi\right)^\intercal \vect(\Gamma),
	\end{align*}
	thus obtaining the same upper bound of equation \ref{eq:GammaQP} and completing the proof. 
\end{proof}

The results of lemma \ref{lemma:optimal_gamma_qp} allow us to partially speed up the solution process, e.g., by deploying optimization methods specifically designed for the solution of nonlinear problems or capable of exploiting their structure \cite{Waechter_2005, KARDOS2022108613}. However, it is important to notice that---if we consider a discrete segmentation of the input data space---we can show that each subproblem solved by the GOAL algorithm admits a closed-form solution, and we really need to carefully weight the trade-off between this restriction and the resulting computational cost. Indeed, while imposing the affiliation of every input data point to a single box in the rotated lower-dimensional gauge does not appear like an extremely binding constraint---as shown also by the good results obtained in section \ref{sec:experiments} for the problems pertaining to bioinformatics and climate science---on the other hand it allows to move the GOAL algorithm from a polynomial to a linear scaling of the computational cost. And the latter feature results particularly important in the solution of small data learning problems, which---despite the limited amount of data samples for the training set---can still deal with feature spaces with hundred of thousands of dimensions, especially in the natural sciences. In theorem \ref{the:GOAL}, we rely on the results of lemma \ref{lemma:optimal_rotation} and \ref{lemma:optimal_gamma_qp} to summarize the main features of the GOAL algorithm, concerning the optimal solution, its monotonic convergence and the scaling of the iteration cost in case of a discrete feature space segmentation.

\begin{theorem}\label{the:GOAL}
	Given the minimization of the objective function $\mathcal{L}_{\text{GOAL}}$ in equation \ref{eq:GOAL}---under the feasibility constraints in \ref{eq:Lambda}, \ref{eq:Gamma} and \ref{eq:R}---the following statements hold:
	\begin{itemize}
		\item [(i)] The optimal segmentation $\Gamma^*$ is piecewise linear in the feature space $X$.
		\item [(ii)] The GOAL algorithm converges monotonically to the approximated solution.
		\item [(iii)] In case of a discrete segmentation of the feature space, i.e., by replacing \ref{eq:Gamma} with:
		\begin{equation}\label{eq:Gamma_discrete}
			\Omega_{\Gamma} := \left\{ \Gamma\in \mathbb{R}^{K \times T} \bigg | \, \forall k,t: \; \Gamma_{k,t} \in \{0,1\}\; \wedge \; \forall t: \; \sum_{k=1}^{K} \Gamma_{k,t}=1 \right\},
		\end{equation}
		every GOAL iterative sub-problem presents a closed-form solution, and the total iteration cost scaling is $\mathcal{O}(T(KD+K+D+KM+M)+D(G^2+KG)+MK$.
	\end{itemize}
\end{theorem}

\begin{proof}
	Proposition (i) can be proved---after fixing the gauge rotation matrix $R$ and the matrix of box coordinates $S$---directly from the application of lemma 14 from \cite{gerber20}, which deals with a similar problem. Proposition (ii) can be proved as follows. According to lemma \ref{lemma:optimal_gamma_qp}, we know that the $\Gamma$-step of the GOAL algorithm can be approximated with its upper bound---consisting in a convex QP problem on a simplex domain of linear constraints---which leads to the monotonic minimization of the objective function $\mathcal{L}_\text{GOAL}$. As far as concerns the $S$-step and the $\Lambda$-step, their monotonic convergence can be derived from the results in theorem 1 from \cite{horenko2020scalable}. Finally, the $R$-step---according to the results in lemma \ref{lemma:optimal_rotation}---presents a closed-form solution, which automatically leads to the monotonic decrease of the objective function when solving this specific subproblem while keeping the other variables fixed. Then, given the results reported above about each subproblem, we can conclude that the GOAL algorithm results in the monotonic minimization of the objective function $\mathcal{L}_\text{GOAL}$. Proposition (iii) can be proved by noting that---under constraint \ref{eq:Gamma_discrete}---we get:
		\begin{equation}\label{eq:GOAL_discrete}
	 		\mathcal{L}_{\text{GOAL}}=\frac{1}{T}\sum_{d,t,k=1}^{D,T,K}\Gamma_{k,t}\left(X_{d,t}-\left\{RS\right\}_{d,k}\right)^2 -\frac{\varepsilon_{\text{CL}}}{TM}\sum_{m,t=1}^{M,T}\Pi_{m,t}\sum_{k=1}^K\Gamma_{k,t}\log\left(\Lambda_{m,k}\right),
		\end{equation}
		where the discrete cluster affiliation coefficient has been brought outside of the Euclidean norm and of the relative entropy since it can only assume either value $0$ or $1$. Given this new formulation of the objective function in equation \ref{eq:GOAL_discrete}, we can deploy the method of Lagrange multipliers and derive a closed-form solution for the $S$-step, the $\Lambda$-step and the $\Gamma$-step---refer, e.g., to \cite{vecchi2022espa+} for the solution of a similar problem. As far as concerns the $R$-step for the computation of the optimal rotation matrix in the lower-dimensional gauge, we know that the results of lemma \ref{lemma:optimal_rotation} are independent on the type of segmentation and, therefore, hold for both a fuzzy $\Gamma$ (as in the general GOAL formulation) or for a discrete $\Gamma$ (as in this specific case). Thus---according to the statements above--- we can conclude that, as claimed, all subproblems solved by the GOAL algorithm admit closed-form solutions. The complexities of the fours subproblems are, respectively, $\mathcal{O}(DT+K(T+DG))$ for the $S$-step, $\mathcal{O}(M(K+T))$ for the $\Lambda$-step, $\mathcal{O}(KT(M+D+1))$ for the $\Gamma$-step and $\mathcal{O}(DG^2)$ for the computation of of the truncated SVD in the $R$-step. By combining these results and by dismissing the constant terms, we can conclude that GOAL achieves an iteration cost scaling of $\mathcal{O}(T(KD+K+D+KM+M)+D(G^2+KG)+MK$, which is linear with respect to the dimensions $D$ and $T$ of the problem input data matrix $X$.
\end{proof}

To conclude the methodological part, in algorithm \ref{alg:GOAL} we report the four subproblems solved during the minimization of the GOAL objective function \ref{eq:GOAL}, together with their closed-form solution in case of a discrete segmentation. This version of the algorithm corresponds to the one used in the experiments reported in the next section.

\begin{algorithm}
\small
\caption{Analytically-computable GOAL algorithm for the minimization of the functional in Eq.~\ref{eq:GOAL} under the feasibility constraints in \ref{eq:Gamma}, \ref{eq:Lambda} and \ref{eq:R}}\label{alg:GOAL}
\SetAlgoLined
\KwData{$X$, $\Pi$ and fixed $K$, $\varepsilon_{CL}$, $\varepsilon_E$, $tol$; random feasible initial values for $\Gamma^{(1)}$ and $R^{(1)}$}
\KwResult{Optimal values of $S$, $\Lambda$, $\Gamma$ and $R$ minimizing the function $\mathcal{L}_{\text{GOAL}}$ in \ref{eq:GOAL}}
 $I=1;\, \mathcal{L}^{(I)}=\infty;\, \Delta \mathcal{L}^{(I)}=\infty$\;
 \While{$\Delta \mathcal{L}^{(I)}>tol$}
 {
 	\underline{\textbf{$\mathbf{S}$-step}}: for fixed $\Gamma^{(I)}$ and $R^{(I)}$, compute the entries $\displaystyle \left\{RS\right\}^{(I)}_{d,k} = \frac{\sum\limits_{t=1}^T \Gamma_{k,t}^{(I)}X_{d,t}}{\sum\limits_{t=1}^T \Gamma_{k,t}^{(I)}}$ and the resulting box centres coordinates $S^{(I)}=(R^{(I)})^\intercal \left(RS\right)^{(I)}$ \;
 	\underline{\textbf{$\mathbf{\Gamma}$-step}}: for every $t$ and fixed $S^{(I)}$, $\Lambda^{(I)}$, $R^{(I)}$, find the discrete affiliations $\displaystyle\Gamma_{:,t}^{(I+1)}$ through the closed-form solution:
    	$	\Gamma_{k,t}^{(I+1)} = \left\{\begin{array}{lr}
        1, & \text{if } k=k',\\
        0, & \text{otherwise},
        \end{array}\right.
    	$
    	where $k'$ is
   		$\displaystyle \argmin_{k'} \left[\sum_{d=1}^D \left(X_{d,t}-\left\{R^{(I)}S^{(I)}\right\}_{d,k'}\right)^2 -\frac{\varepsilon_{CL}}{M} \sum_{m=1}^M\Pi_{m,t} \max\left[\log \left(\Lambda_{m,k'}^{(I)}\right),tol\right]\right]$\;
	\underline{\textbf{$\mathbf{\Lambda}$-step}}: for fixed $\Gamma^{(I+1)}$, compute analytically the conditional probabilities in $\Lambda^{(I)}$, through the closed-form solution $\displaystyle\Lambda_{m,k}^{(I)} = \frac{\left\{\Pi(\Gamma^{(I+1)})^\intercal\right\}_{m,k}}{\sum\limits_{m=1}^M\left\{\Pi(\Gamma^{(I+1)})^\intercal\right\}_{m,k}}$\;
	\underline{\textbf{$\mathbf{R}$-step}}: for fixed $\Gamma^{(I+1)}$ and $S^{(I)}$, compute the gauge-optimal rotation $\displaystyle R^{(I+1)} = U\widetilde{I}V^\intercal $, with $U$ and $V$ obtained from the SVD of $X(\Gamma^{(I+1)})^\intercal (S^{(I)})^\intercal$ \;
	\underline{\textbf{Update step}}:	compute $\mathcal{L}^{(I+1)}= \mathcal{L}(S^{(I)},\Lambda^{(I)},\Gamma^{(I+1)},R^{(I+1)})$ as $\mathcal{L}_{\text{GOAL}}$ in \ref{eq:GOAL}\;
	$I = I + 1$\;
	$\Delta L^{(I)}= L^{(I-1)}-L^{(I)}$\; 
 }
\end{algorithm}


\section{Experimental Results and Numerical Comparisons}\label{sec:experiments}

In this section, we benchmark the proposed GOAL algorithm---both in terms of learning performance and computational cost---against several other state-of-the-art competitors from ML and DL for the solution of both synthetic and real-world problems in the small data regime. The code used to perform the simulations has been written entirely in MATLAB (version: R2022a) in order to eliminate eventual biases induced by the choice of different programming languages and to provide a fair comparison of the different methods. The experiments have been run on two different architectures, depending on the size of the problem: the applications to the synthetic and climate science data have been performed on a laptop with an Intel Core i7-8665U CPU 1.90GHz (4 cores) processor with 16 GB RAM, while the gene-inference problem has been run remotely on a machine with 2 CPUs (Intel Xeon Gold 6240R 2.4G, 24C/48T) and 384 GB RAM (DDR4-2933) using up to 98\% of the 48 physical cores and ${\sim}$60GB memory.


\subsection{Methodological Details and Description of the Data Sets}\label{sec:methods}

In this section, we briefly outline the characteristics of the synthetic and real-world data sets used in the simulations, and we provide some details about the methods used in the experiments and their implementation as well as on the hyperparameter tuning phase.

\paragraph{Description of the Experimental Data Sets.} The experiments have been performed on three different data sets dealing, respectively, with: (i) bioinformatics-inspired synthetic data; (ii) real-world data pertaining to a practical climate science forecasting application (prediction of the warm phase of the El Ni\~no phenomenon); (iii) real-world genomics data aimed at giving insight into the gene-inference prediction problem.

Data set (i) is derived from a real small data problem in bioinformatics, i.e., the prediction of presence or absence of a longer lifespan in some worms of the species \emph{Caenorhabditis elegans}. Specifically, the main task consists in the solution of the binary classification problem in presence of only two relevant dimensions---which would correspond to the two genes that, in a real-world setting, would determine the presence of a longer lifespan through their co-expression---while the other dimensions of the feature space are just noise with no relevant information. The reason behind the construction of synthetic data from a real-world example stems from the possibility of freely tuning the number of irrelevant dimensions for benchmarking purposes: as shown by \cite{horenko2020scalable} and \cite{vecchi2022espa+}, increasing the number of features shifts the problem towards the small data regime, and allows a clear assessment of each method capability to discriminate the information relevant for the classification task.

Data set (ii) deals with an example from climate science and consists in the prediction of the active phase of the El Ni\~{n}o--Southern Oscillation (ENSO)---i.e., a meteorological phenomenon with a highly irregular periodicity described by a variation of wind and sea surface temperatures in the Pacific Ocean. Given its strong global impact \cite{cook2017impact, gupta2020impact}, the prediction of El Ni\~{n}o currently received increasing attention in the DL community \cite{10.3389/fphy.2019.00153, horenko2023oncheap}, but the performance of the available reported tools deteriorates quickly beyond 12-16 months lead time \cite{ham2019deep, he2019dlenso}. In this letter, we consider the ENSO prediction over different time horizons (6-, 9-, 12- and 24-months-ahead, respectively), by using data from the Nino3.4 index in the period 1950-2007 (the warm phase is encoded as a binary variable equal to 1 if the index is above the standard value of 0.4). As features, we rely on a series of deseasonalized climate proxies derived from the sea surface temperature and the ocean surface layer depth \cite{an2020fokker} over four different time horizons, consisting---specifically---in time $t$, $t-1$, $t-2$ and $t-3$, with the integers representing the numbers of months. The values of the two variables considered are obtained from two discretization grids with different resolutions: the sea surface temperature (SST) considers $180$ evenly spaced latitudinal points and $360$ longitudinal points---thus resulting in a $180\times 360$ grid---while the ocean surface layer depth ($\Delta$Z) relies on a total of $50$ latitudinal points and $180$ longitudinal points---thus resulting in a $50\times 180$ grid. In total, this example from climate science considers $T=720$ data instances coming from the Nino3.4 index and $D=295\,200$ explanatory features and---due to the significant discrepancy between the number of dimensions and the number of observations---represents an extremely challenging small data learning problem. Thus, it allows us to test how much the methods are able to reduce the dimensionality of the learning space and to dismiss the features irrelevant for the binary classification task. 

Data set (iii) considers instead a real-world small data problem from bioinformatics, by dealing with the prediction of a single gene activity starting from the activity profile of other co-expressed genes. Specifically, in order to assemble the gene network data set, we used experimental data from the ENCODE (encyclopedia of DNA elements) data portal \cite{davis2018encyclopedia} for 185 ChIP-seq (chromatin immunoprecipitation followed by high-throughput sequencing) data sets describing the interaction profiles of the histone mark H3K4me3 in various biosamples (tissues or cell types). To increase the quality and reliability of the experimental data, we selected only replicated data sets with the ENCODE status \emph{released}, and we excluded all data sets with any biosample treatment or genetical modification. Eventually, only fully preprocessed data was taken into account---i.e., the data sets for which the \emph{replicated peaks} for the genome assembly GRCh38 (\emph{Homo Sapiens}) was available, created by the ENCODE histone ChIP-seq preprocessing pipeline. This final outcome considered represents the experimental ChIP-seq profile, which contains the genomic regions in which an interaction was found to be significant. After applying all these criteria, we downloaded 185 files, representing the number of instances of the response variable---i.e., the statistics size $T$. The histone mark H3K4me3 was selected because of its strong association to active promoters and due to its major impact on the regulation of gene expression. According to what we discussed above, we defined a gene to be either active (1) or not active (0) for a given ChIP-seq profile, depending on whether the latter contained a H3K4me3 interaction within the genes promoter region or not. The gene annotation procedure was applied to the \emph{replicated peaks} files using the \emph{annotatePeaks} function from the homer framework \cite{heinz2010simple}. The final gene-activity network data set was restricted to a subset of 263 genes, selected based on the KEGG entries for the two biggest metabolic pathways---i.e., 133 genes for the Oxidative phosphorylation pathway (hsa00190) and 130 genes for the Purine metabolism pathway (hsa00230) \cite{ogata1999kegg, kanehisa2019toward, kanehisa2021kegg}. In this way, the network properties were induced to the data set for a large amount of genes, but---on the other hand---the biosample is expected to present a lower degree of vairation, since these metabolic pathways are relevant for any kind of tissue or cell type. Among these 263 genes, 230 are active in at least one interaction profile, while 2 are active in all profiles and, thus, not informative. Since the regulatory patterns induced by the networks can be sparse, also rarely active genes can be useful for training a model and---thus---we decided to include them, resulting in $D=227$ features to perform the various classification and inference tasks. 

\paragraph{Details on the simulation benchmark.} We report here a brief description of the ML methods that have been compared against the proposed GOAL algorithm in the experimental part, as well as some eventual details on their MATLAB implementation.
\begin{itemize}
	\item A Gaussian mixture model (GMM) is a multivariate distribution consisting of $k$ multivariate Gaussian distribution components, each one defined by its mean and covariance \cite{peel2000finite}. The MATLAB \emph{Statistics and Machine Learning Toolbox} provides the function \texttt{fitgmdist()} which---given as input a numeric data matrix $X$ and a number of components $k$---returns a GMM distribution model fit on the input data. The obtained distribution has then been used to train: (i) a Bayesian classifier \cite{hastie2009elements} with the function \texttt{fitcnb()} by passing it, along with the training data, the target labels; (ii) a neural network (NN) \cite{glorot2010understanding}, where the weights have been adjusted with the propagation of the data through a sequence of hidden layers.
	\item K-means---a widely known clustering technique which partitions the data into an arbitrary number $k$ of clusters by minimizing the intra-cluster variance \cite{arthur2007k}---has been used as a preprocessing step to foster the classification performed by a Bayesian classifier and a NN, as we did in the case of GMM. The chosen implementation relies on the function \texttt{kmeans()}, contained in MATLAB \emph{Statistics and Machine Learning Toolbox}.
	\item For the sake of comparison, a NN has also been trained: (i) without any preprocessing; (ii) with the aid of a PCA routine \cite{jolliffe2003principal} maximizing the variance of a linear combination of the original features; (iii) following a multi-layer deep learning architecture. 
	\item Support vector machines (SVM) \cite{cristianini2000introduction} is a supervised learning method projecting the data points onto a higher-dimensional space and finding the hyperplane separating the two classes. We used the function \texttt{fitcsvm()}---provided by the MATLAB \emph{Statistics and Machine Learning Toolbox}---and the hyperparameters that have been tuned for this method are the misclassification cost and the kernel mapping the data points into the new space.
	\item A generalized linear model (GLM) \cite{mcfadden1973conditional} is a nonlinear model which allows the response variable to have a distribution different than normal. A sparse GLM has been trained through the function \texttt{fitglm()}, provided by the MATLAB \emph{Statistics and Machine Learning Toolbox}.
	\item Linear discriminant analysis (LDA) \cite{fisher1936use} assumes instead that each class has the same covariance matrix, but different means. A LDA classifier is trained with the function \texttt{fitcdiscr()}---provided by the MATLAB \emph{Statistics and Machine Learning Toolbox}---by toggling the \emph{DiscrimType} to ``linear''.
	\item Long short-term memory (LSTM) is an artificial recurring neural network which has been chosen as the most prominent representative of DL \cite{hochreiter1997long}. The MATLAB implementation used in the experiments is contained within the \emph{Deep Learning Toolbox} and the solution of the stochastic minimization problems arising within the method has been performed through the ADAM algorithm \cite{kingma2014adam}. During the simulations, we considered as a tunable hyperparameter the number of hidden layers, while the number of fully connected components in the fully connected hidden layer has been chosen equal to 2---i.e., the number of classes in this binary classification task.
	\item Random decision forests (RF) is a classification algorithm which performs the class labelling according to several decision tree \cite{breiman2001random, breiman1984classification}. It has been included within the analysis of data set (iii), since this method is widely used in bioinformatics to solve classification tasks. The chosen implementation is based on the MATLAB function \texttt{TreeBagger()}, which is contained within the \emph{Statistics and Machine Learning Toolbox}. Further details about the hyperparameters used in the simulations are provided in Section \ref{sec:result_discussion}, in the discussion of the experimental results.
	\item Finally---as already mentioned---eSPA is a classification algorithm originally developed in \cite{horenko2020scalable}, which simultaneously addresses the optimal discretization, feature selection and supervised classification problems, and it is particularly suited to operate in the small data learning regime. In this comparison, we rely on the improved algorithm provided in \cite{vecchi2022espa+}, which presents a significantly lower computational cost due to the assumption of a discrete segmentation of the feature space, which results in a closed-form solution for each optimization problem solved by the method. The implementation used in the benchmark is written entirely in MATLAB and the exploration of the hyperparameter space (i.e., number of discretization boxes $K$, regularization of the entropy maximization $\varepsilon_\text{E}$ and regularization of the importance of the supervised classification problem in the overall classification task $\varepsilon_\text{CL}$) is parallelized directly through the \emph{Parallel Computing Toolbox}. 
 \end{itemize}

\subsection{Discussion of the Experimental Results}\label{sec:result_discussion}

\begin{figure}[h!]
\begin{center} 
    \includegraphics[width=0.95\textwidth]{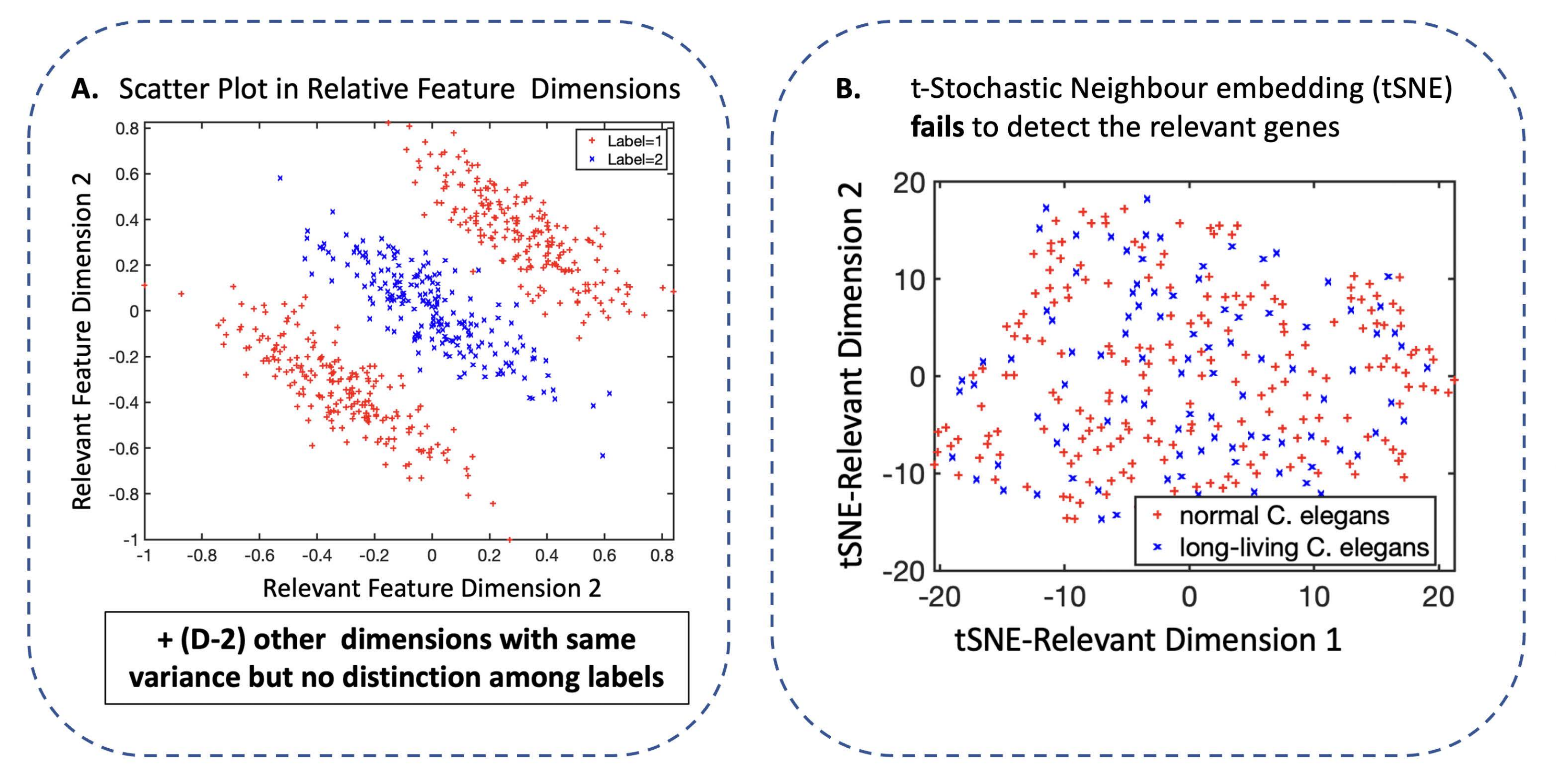}  
 \end{center}
   \caption{Visualization of the bioinformatics synthetic data set (i).}
    \label{fig:1res}
\end{figure}

In this section, we present the results of the benchmarking of GOAL against the state-of-the-art ML competitors described in section \ref{sec:methods}. In panel A of figure \ref{fig:1res}, we report the graphical representation of the bioinformatics data set (i) provided in \cite{vecchi2022espa+}, and visualize the learning problem in the two dimensions which are relevant for the classification task. Specifically, the long-living worms are represented by the single group of blue elements in the middle of the figure, while the red elements correspond to the normal-living worms. It is important to notice that the two groups are not linearly separable and that in the benchmark we consider other $D-2$ dimensions which bear no relevant information for the classification task, since the two groups are not distinguishable if we try to label them according to different features than the first two.

\begin{figure}[h!]
\begin{center} 
    \includegraphics[width=0.95\textwidth]{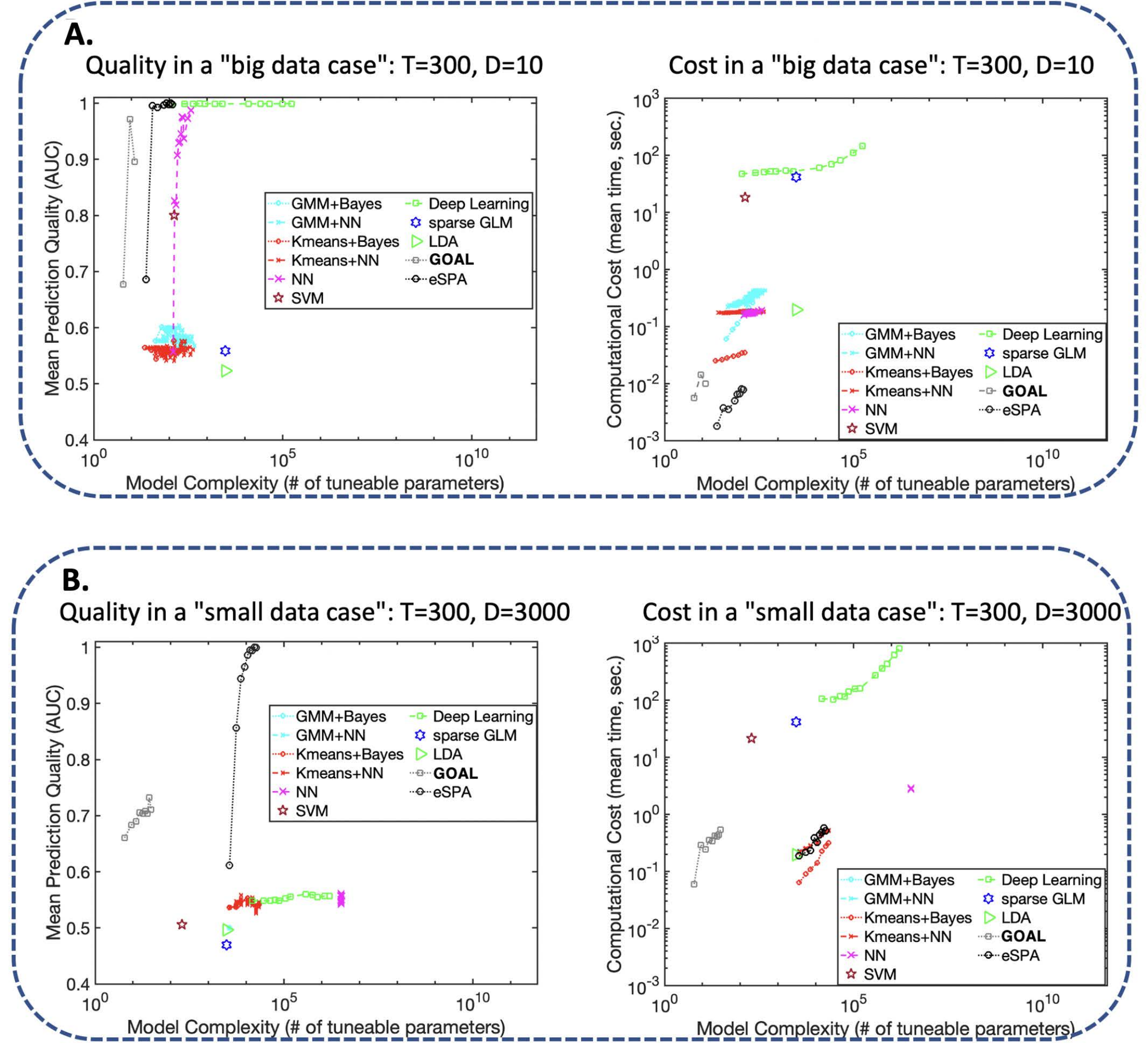}  
 \end{center}
   \caption{Comparison of GOAL against common ML tools on dataset (i).}
    \label{fig:2res}
\end{figure}

In panel B of figure \ref{fig:1res}, we show that the identification of the dimensions relevant for the classification task is particularly challenging even for state-of-the-art methods, like the t-Stochastic Neighbor Embedding (tSNE) algorithm, which is a dimensionality reduction method for embedding and visualizing a higher-dimensional space into a lower-dimensional space---i.e., 2D in this case \cite{van2008visualizing}. In particular, unlike the visualization that we would obtain by considering the first two genes---i.e., the only two which are relevant for the classification task---we can notice that the output provided by tSNE consists in the perfect overlapping of the two groups, which cannot then be separated. In figure \ref{fig:2res}, we show instead the results of the comparison of the different ML algorithms on the classification task (i), by reporting the outcome for for the big data (i.e., $T\gg D$, with $T=300$ and $D=10$) case in panel A and for the small data (i.e., $T\ll D$, with $T=300$ and $D=1000$) case in panel B, as well as the model complexity on the $x$-axis---which is measured in terms of the number of tunable parameters in each of the different algorithms.

The model quality is measured in terms of Area Under the ROC curve (AUC), and---as we can notice in panel A---the data set is roughly balanced, with the majority class (i.e., normal-living worms) accounting for two thirds of the overall training and test instances. The performance values reported in the figure do not come from a single instance, but---in order to foster their statistical significance---represent the average of 100 cross-validations, with the same training/validation/test splits for all the methods. As we can notice, the eSPA+ algorithm achieves the best performance both in the big and small data cases, while requiring a significantly smaller number of tunable parameters when compared with the other methods, even at the same level of performance. On the other hand, the GOAL method proposed in this paper is able to achieve an AUC of roughly $0.95$ with around 10 tunable hyperparameters in the big data case and an AUC or around $0.7$ with the same number of hyperparameters in the small data case. These results, while being better than the ones achieved by the other methods---many of which achieved an AUC of $0.5$, representing their inability to discriminate between the normal-living and the long-living worms---are still worse than the ones achieved by eSPA+. However, this behaviour was expected, since GOAL reduces the dimensionality of the feature space by performing a rotation in a lower-dimensional gauge but---at the moment---does not possess any metric (like, e.g., the entropy maximization performed by eSPA+) which allows it to understand that the sparse classification rule lies in the dimensions with the smallest variance. Indeed, as we can observe in figure \ref{fig:1res}, the variance is significantly smaller in the first two dimensions---where the worm clusters are easily identifiable---and larger in all the other dimensions, thus posing a real challenge for those methods which perform a reduction of the feature space dimensionality by using this information. However, we can still notice that the GOAL algorithm presents a very low computational cost---in the order of $10^{-2}$ for the big data case and $10^{-1}$ for the small data case---while, e.g., LSTM is 4 orders of magnitude slower, and fails completely to perform the binary classification in the small data case.

\begin{figure}[h!]
	\centering
	\includegraphics[width=\textwidth]{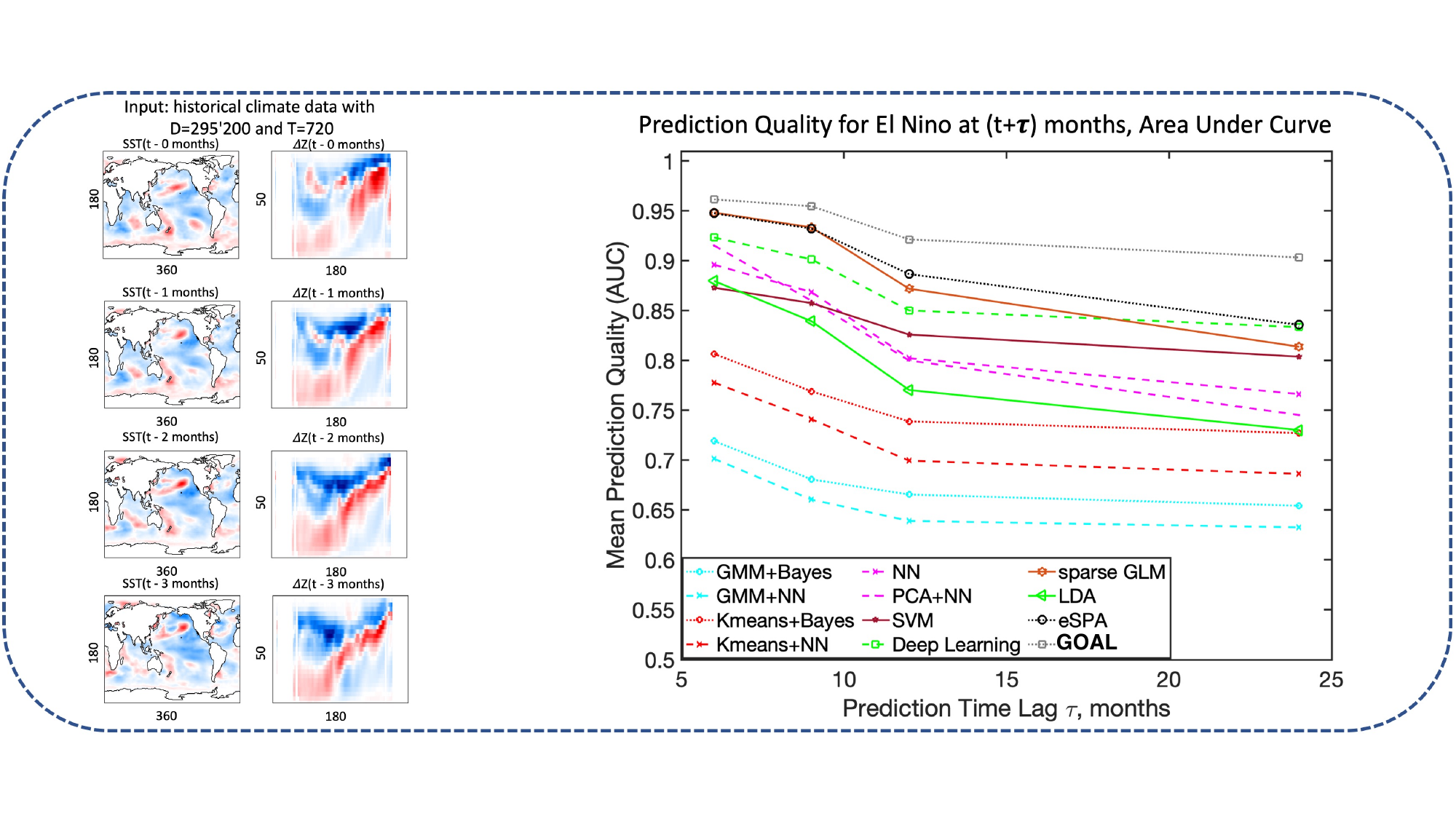}
	\caption{Comparison of GOAL against common ML methods on the El Ni\~no data set.}
	\label{fig:3res}
\end{figure}

While the results obtained for GOAL in the synthetic bioinformatics example are better than the ones achieved by the majority of its competitors, they would not still justify employing this method instead of eSPA+ for the solution of small data classification tasks. Thus, in order to show the full potential of the method proposed in this letter, in figure \ref{fig:3res} we consider the application of GOAL and of the other ML algorithms to the prediction of El Ni\~no phenomenon contained in data set (ii) at different time horizons. 

Specifically, we can notice---in the left part of figure \ref{fig:3res}---a graphical representation of this challenging learning problem (with $D=295\,200$ dimensions and $T=720$ data instances), while in the right part we report the results obtained for a \mbox{6-month-,} \mbox{9-month-,} 12-month- and 24-month-ahead prediction of the El Ni\~no phenomenon, with the last time horizon resulting particularly challenging. In the case of the El Ni\~no prediction, we were not able to perform multiple cross-validation, but we were forced to consider a single training/test split, since the data had to be temporally separated. This resulted in a roughly balanced training and test sets, with almost the same percentage of instances of the majority class---i.e., the presence of the El Ni\~no phenomenon after $\tau$ months---namely 33\% and 28\%, respectively. The results show---as expected---a decrease in the performance of all algorithms from the initial value obtained in the 6-months-ahead prediction, and all methods are able to achieve at least a AUC of $0.65$, which is better than the one obtained in data set (i). However, we can notice that---in this example---GOAL dominates its competitors, especially in the case of the 24-month-ahead prediction, where it achieves an AUC of roughly $0.9$ at a fraction of the computational cost. Indeed---as explained also in section \ref{sec:GOAL}---the rotation and embedding of such a high-dimensional space in a low-dimensional gauge results particularly beneficial and significantly aids the solution of the classification task.

In order to complete the benchmarking of GOAL against the other ML methods, we need to consider the results obtained from the bioinformatics example in data set (iii). Specifically, for the gene network inference tasks, the activity values of the genes were used to predict the activity values of a certain gene used to define the class feature. As some genes are rarely active or active in almost all instances, not every gene was used to define the class feature. Only genes with at least 5\% or less than 95\% activity (i.e., number of positive or simultaneously active genes) were used to create a class feature, resulting in 182 classification cases, where each of the other genes was used as predictive feature. Concerning the data set balance, we can notice that it varies significantly depending on the gene considered, but it presents a trend towards 20\%-30\% number of instances in the majority class (i.e., the gene is activated), with some perfectly balanced cases as well. The results---in terms of both performance metrics (i.e., AUC and accuracy) and computational cost---are summarized in table \ref{tab:1}. It is worth noticing that, in this last case, we were able to perform only 5 cross-validations with random training/test splits since the problem solution was computationally expensive, especially for methods like LSTM and RF which do not rely on closed-form solutions.

\begin{table}[h]
  \centering
  \begin{tabular}{p{0.16\textwidth}|>{\centering}p{0.175\textwidth}>{\centering}p{0.185\textwidth}>{\centering}p{0.175\textwidth}>{\centering\arraybackslash}p{0.175\textwidth}}
    \toprule
      Metric     & GOAL  & eSPA  & RF    & LSTM       \\
\hline \hline 
		AUC & \bf{0.910} & 0.879 & 0.801 & 0.743 \\ 
		ACC & \bf{0.828} & 0.826 & 0.797 & 0.771 \\ 
		time(s) & 0.019 & \bf{0.011} & 9.840 & 39.878 \\ 
		\# parameters & 5154 & \bf{5064} & * & 44826 \\ 
		\hline \hline 
		$\Delta$AUC, C.I. & $\bf{-0.003\pm0.001}$ & $-0.033\pm0.002$ & $-0.112\pm0.010$ & $-0.170\pm0.008$ \\ 
		$\Delta$ACC, C.I. & $\bf{-0.024\pm0.003}$ & $-0.027\pm0.004$ & $-0.055\pm0.005$ & $-0.081\pm0.007$ \\ 
		\bottomrule
  \end{tabular}  \caption{Benchmarking GOAL against eSPA, RF and DL for the inference of the gene-activity networks contained in data set (iii). $\Delta$AUC and $\Delta$ACC represent the distance and uncertainty of deviation from the best performing model for AUC and ACC.\medskip}\label{tab:1}
\end{table}

As we can notice in table \ref{tab:1}, we restricted our attention only to GOAL and two of its previous competitors---namely eSPA and LSTM---while we decided to include also RF due to its successful application in many genomics-related problems. For each case, the four algorithms were run on 5 cross-validations with random training/test splits (with 75\% of the instances for training and 25\% for test). The results reported in the table are the mean AUC and ACC obtained across all 182 genes and cross-validations, as well as the difference from the measure achieved by the best performing model together with the confidence intervals computed by 1.96 times the standard error of the mean according to all genes and cross-validations (indicated as $\Delta$AUC and $\Delta$ACC in the table). As a reference, we report also the runtime (in seconds) of the competing algorithm and the number of parameters included in the best-performing model. As we can observe, GOAL outperforms its competitors both in terms of AUC and ACC, while it presents also the smallest difference from the best-performing model---especially in terms of AUC---in all different scenarios. Since table \ref{tab:1} provides an aggregate representation of the results, it is worth mentioning that, out of the 910 classification scenarios considered---i.e., 182 genes, each with 5 cross-validations---GOAL achieved the best AUC in 82.7\% of the cases, followed by eSPA (16.3\% of the cases), while the performance of RF and LSTM was almost negligible in all instances (i.e., they were the best model only in 0.9\% and 0.1\% of the cases, respectively).

The particularly dissatisfying performance of RF prompted us to perform additional simulations and to briefly evaluate if the results obtained for this method were actually caused by a bad choice of the hyperparameter grid. In the simulations summarized in table \ref{tab:1}, we determined the choice of the hyperparameters according to some preliminary test performed on subsets of data set (iii), and we considered a rather high number of trees---namely $100$, $500$ or $1000$ trees---in order to achieve a high predictive performance. We used either the Gini-Index or Entropy as splitting criteria and a minimum number of observations per leaf of $1$ and $3$ (with $1$ representing the default value suggested by the MATLAB implementation). In order to assess the relevance of the hyperparameter choice on RF performance on the gene-network inference task, we extended the grid search as follows: the \emph{maximum number of decision splits} has been considered with both the default value $n-1$ and with $\frac{n-1}{2}$, where $n$ is the number of samples in the training set; the values for minimum leaf size were changed to $1, 5, 10, 20$; the number of trees was extended to $5, 10, 25, 50, 100, 200, 500, 1000$; the splitting criteria were left unchanged. The main difference between the two configurations lies---in particular---in the fact that, in the original analysis, we did not consider very small tree sizes---i.e., values below $50$ trees. After incorporating these changes, the AUC significantly improved, reaching a value of $0.859$, while the runtime decreased, thus providing evidence towards the fact that also models with less decision trees have a chance to perform well. However, we also observed that the accuracy decreased significantly to $0.724$, and the reason behind this apparently odd behaviour is that the RF model selected as best-performing often contains only $5$ or $10$ trees. Such small models are fast, but also strongly limited in describing the classification probabilities: for example, a model with $5$ trees is not able to provide more than $6$ different probability scores for all validation samples. While for a binary classification scenario this might not have a negative impact on the AUC, it can still drastically change the observed accuracy, and---by using very small numbers of trees---one might face issues related to less stable classification probability scores, as we exactly observed in the drop in accuracy observed for the gene-network inference data set. We can then conclude that even a change in the hyperparameter choice for RF is not going to make this model more performing than GOAL, thus confirming the results presented in table \ref{tab:1}.


\section{Conclusions and Future Work}\label{sec:conclusions}

In this letter, we proposed the GOAL algorithm, which starts from the idea of optimal gauge-rotation to simultaneously solve the features dimension reduction, feature segmentation and the KL-divergence-optimized labelled classification learning problems. The novelty of the algorithm lies in the fact that it provides a joint solution to all the problems discussed above, unlike the other methods currently available: for example, the singular value decomposition (SVD) allows to perform the feature reduction in a partially similar way, but SVD cannot simultaneously solve the discretization and the labelled data learning problems while solving the dimension reduction and gauge-rotation problems, and---certainly---it does not allow to achieve this solution in a linearly scalable and regular way as the GOAL algorithm. From a more technical perspective, we have proved that the optimal rotation matrix in the lower-dimensional gauge can be computed by means of a closed-form solution which is independent on the type of discretization (i.e., fuzzy or discrete) chosen for the input data, and that the optimization problem solved by the algorithm can be approximated through a convex QP problem in case of a fuzzy segmentation of the feature space. These two results allow an advantageous reduction of the overall computational complexity of the algorithm, which results significantly faster than the majority of its competitors, while achieving a higher classification performance. The superiority of the GOAL algorithm has been confirmed also by several benchmarking experiments, which have been performed both on synthetic examples pertaining to the small data regime, as well as on real-world problems from climate science and bioinformatics. In particular, the results obtained for the 24-month-ahead prediction of El Ni\~no warm phase were better than the ones achieved by its competitors, despite the challenging discretization of a feature space containing $295\,200$ dimensions, while the solution of the gene-network inference problem highlighted the stability of the algorithm even in the solution of problems characterized by a different degree of imbalance between the majority and the minority class. However, despite the promising results obtained in the simulations, the work discussed in this letter presents also several limitations. For example, we still need to clearly assess the impact of the feature space segmentation on the overall algorithm performance since, in all the simulations, we assumed a discrete segmentation of the input data in the lower-dimensional gauge. While considering a fuzzy segmentation (i.e., values of $\Gamma\in[0,1]$) could potentially improve even further the classification performance, it could also definitely have an impact on the algorithm computational cost---which would be significantly higher due to the lack of closed-form solutions for the various optimization steps solved by the algorithm. Finally, a promising future research direction could consist in the improvement of the solution of the eigenvalue problem required---at each iteration---to determine the optimal rotation matrix in the lower-dimensional gauge, since this problem currently represents the most expensive computational task of GOAL.

\bibliographystyle{unsrt}
\bibliography{refs}

\end{document}